\documentclass{article}

    \PassOptionsToPackage{numbers, compress}{natbib}
\usepackage[final]{neurips_2021}

\usepackage[utf8]{inputenc} % allow utf-8 input
\usepackage[T1]{fontenc}    % use 8-bit T1 fonts
\usepackage{hyperref}       % hyperlinks
\usepackage{url}            % simple URL typesetting
\usepackage{booktabs}       % professional-quality tables
\usepackage{amsfonts}       % blackboard math symbols
\usepackage{nicefrac}       % compact symbols for 1/2, etc.
\usepackage{microtype}      % microtypography
\usepackage{graphicx} 
\usepackage{color}
\usepackage{xcolor}
\usepackage{makecell}
\usepackage{diagbox} 
\usepackage{multirow}
\usepackage{subfig}
\usepackage{diagbox} 
\usepackage{wrapfig}
\usepackage{pifont}

\usepackage{braket}

\usepackage{amsmath, amsthm}
\theoremstyle{plain}

\theoremstyle{remark}

\usepackage{enumitem}

\usepackage{algpseudocode}% http://ctan.org/pkg/algorithmicx
\usepackage[ruled,vlined]{algorithm2e}


\newcommand{\ts}{\textsuperscript}

\PassOptionsToPackage{hyphens}{url}\usepackage{hyperref}
\hypersetup{colorlinks=true}

\definecolor{Tianlong_color}{rgb}{0.858, 0.188, 0.478}

\title{You are caught stealing my winning lottery ticket! \\ Making a lottery ticket claim its ownership}

\author{%
  Xuxi Chen\textsuperscript{1*}, Tianlong Chen\textsuperscript{1*}, Zhenyu Zhang\textsuperscript{2}, Zhangyang Wang\textsuperscript{1} \\
  \textsuperscript{1}University of Texas at Austin, \textsuperscript{2}University of Science and Technology of China \\
  \small{\texttt{\{xxchen,tianlong.chen,atlaswang\}@utexas.edu,zzy19969@mail.ustc.edu.cn}}
}

\begin{document}

\maketitle

\begin{abstract}
% \textcolor{red}{Use the term ``Lottery-ticket theft" and ``Lottery verification"}
Despite tremendous success in many application scenarios, the training and inference costs of using deep learning are also rapidly increasing over time. The lottery ticket hypothesis (LTH) emerges as a promising framework to leverage a special sparse subnetwork (i.e., \textit{winning ticket}) instead of a full model for both training and inference, that can lower both costs without sacrificing the performance. The main resource bottleneck of LTH is however the extraordinary cost to find the sparse mask of the winning ticket. That makes the found winning ticket become a valuable asset to the owners, highlighting the necessity of protecting its copyright. Our setting adds a new dimension to the recently soaring interest in protecting against the intellectual property (IP) infringement of deep models and verifying their ownerships, since they take owners' massive/unique resources to develop or train. While existing methods explored encrypted weights or predictions, we investigate a unique way to leverage sparse topological information to perform \textit{lottery verification}, by developing several graph-based signatures that can be embedded as credentials. By further combining trigger set-based methods, our proposal can work in both white-box and black-box verification scenarios. Through extensive experiments, we demonstrate the effectiveness of lottery verification in diverse models (ResNet-20, ResNet-18, ResNet-50) on CIFAR-10 and CIFAR-100. Specifically, our verification is shown to be robust to removal attacks such as model fine-tuning and pruning, as well as several ambiguity attacks. Our codes are available at \url{https://github.com/VITA-Group/NO-stealing-LTH}.

%advocates that one can use high-quality subnetworks (i.e., \textit{winning tickets}) to reduce both costs without sacrificing the performance while finding such sparse models still need extraordinary efforts. Meanwhile, it is known that deep learning models face serious intellectual property (IP) infringement threats, and the above intriguing winning tickets are at stake. Many attempts have been made to address these threats, but the existing solutions mainly stem from a perspective of encrypted weights or predictions. 

%However, the sparse models naturally contain benign properties that might be useful for ownership verification, which has never been explored before. To this end, we for the first time propose a brand new topology-based algorithm to verify the ownership of precious lottery tickets, which is termed as \textit{lottery verification}. Specifically, we utilize winning tickets' unique properties to embed signatures as a ``QR code" into their sparse structure. Meanwhile, graph measures are adopted to mark crucial structured paths in sparse tickets as genuine passports for verification. 

%Specifically, when fine-tuning the model the performance will drop XX\%-XX\%, and pruning the model will lead a XX\%-XX\% performance drop. 

\end{abstract}

\renewcommand{\thefootnote}{\fnsymbol{footnote}}
\footnotetext[1]{Equal Contribution.}
\renewcommand{\thefootnote}{\arabic{footnote}}

\vspace{-2mm}
\section{Introduction}
\vspace{-2mm}
Deep neural networks (DNNs) have dramatically raised the state-of-the-art performance in various fields. However, the over-parameterization of DNNs becomes a non-negligible problem. The amount of parameters now is often on the billion scale, which significantly increases the inference cost when using these models. An emerging field of \textit{lottery ticket hypothesis} (LTH) explores a new scheme for pruning the model without sacrificing performance. The core idea is to identify the sparsity pattern ahead of training (or in its early stage) and train a sparse network from scratch. It has been hypothesized~\cite{frankle2018lottery} that DNNs contain sparse networks named \textit{winning tickets} that can be trained to match the test accuracy of the full model. These winning tickets hence have comparable or even better inference performance while potentially reducing the computational footprints.% making them a promising solution for efficient machine learning. 

However, finding winning tickets is a non-trivial task: it involves the training-prune-retraining cycle for several times~\cite{frankle2018lottery}, which is burdensome and computation-consuming. Although other works~\cite{lee2018snip,wang2019picking,You:2019tz} have shown that sparsity might emerge at the initialization or at the early stage of training, the iterative magnitude pruning (IMP) still outperforms these alternatives by clear margins~\cite{frankle2020pruning}. Yet, the powerful IMP method requires multiple rounds of train-prune-train process on the original training set, which is even much more expensive than training a dense network. That makes a found winning ticket a valuable asset to the owners, highlighting the necessity of protecting the winning tickets' copyright.

\begin{wrapfigure}{r}{0.4\linewidth}
\vspace{-1mm}
\includegraphics[width=1\linewidth]{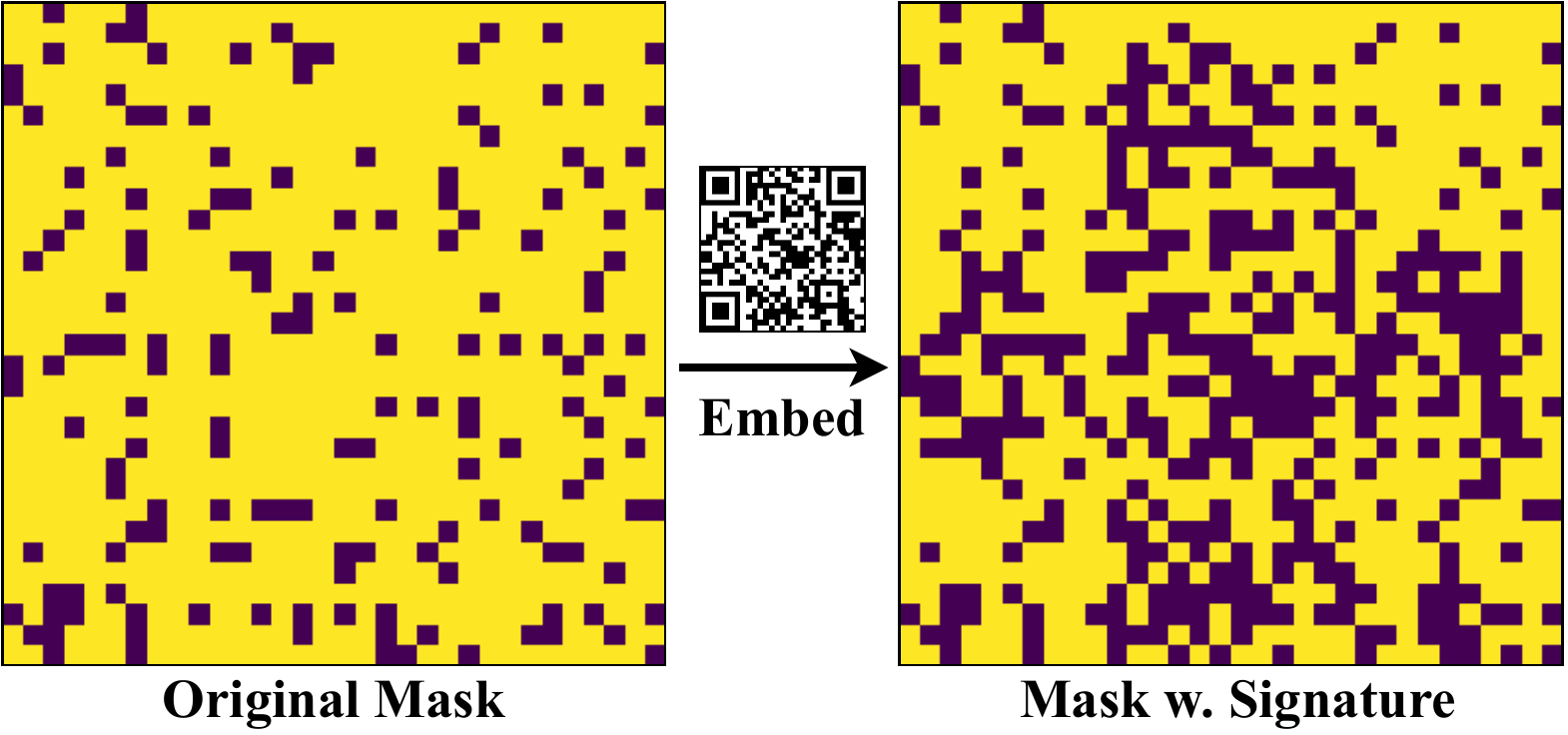}
\vspace{-5mm}
\caption{\small Illustration of embedding signatures into the original sparse mask. These visualizations are projected from $4$D tensors. Dark entries are pruned elements. Note that the actual sparsity of the subnetwork is \textbf{unchanged} after encoding credentials.}
\vspace{-4mm}
\label{fig:tesear}
\end{wrapfigure}

Previous works~\cite{uchida2017embedding,adi2018turning,zhang2018protecting,ma2021undistillable} have shown that deep networks are vulnerable to intellectual property (IP) infringement. For example, one can use transfer learning to adapt a trained model onto a new task or use model compression techniques to create a new sparse model based on the target model. Fortunately, in recent years the ownership verification problem has been addressed with a number of solutions proposed. The key idea is to embed verifiable information, or called \textit{signatures}, into models' weights~\cite{uchida2017embedding,darvish2019deepsigns,zhang2020passport} or predictions~\cite{adi2018turning} without visibly affecting the original performance. By extracting the embedded information from models, one can verify the ownership of models and hence protect their IPs. For the methods that embed information in weights, additional weights regularizers are often used to enforce certain patterns, such as signs. As for the prediction methods, a special training set, which is often called a \textit{trigger} set, is used as additional training data. The model trained upon both the original data and the trigger set can generate desired prediction labels for the privately-held trigger set, while preserving the performance on the original training set. However, those general methods did not take any structural property(e.g., sparsity) into account, leaving chance for improving their gains in the winning ticket mask protection.

%which might be valuable in specific scenarios but never been explored, and we would try to fill this literature gap.

On protecting the IP of winning tickets, we investigate a novel way to leverage \textbf{sparse structural information} for ownership verification (Fig.~\ref{fig:tesear}). This structural information embedded in winning tickets is a good "credential" for ownership verification since the winning ticket at extreme sparsity is naturally robust to fine-tuning and (further) pruning attacks. The winning ticket at extreme sparsity cannot be pruned further; otherwise, the inference performance will drop (hence losing its "value"). Meanwhile, fine-tuning the winning tickets can only tune the weights, but the sparsity pattern will not be changed. However, there remain some key questions to answer: \textit{How to formulate the ownership verification process under the context of the lottery ticket hypothesis? What kind of structural information should be used? How to inject user-specific information into the structure of winning tickets?}  We present answers to these questions in this paper. We summarize our findings as follows:
\vspace{-1mm}
\begin{itemize}[leftmargin=*]
    \item We formulate the lottery verification problem and define two different protection scenarios. We show that even without specific protection, the extremely sparse winning ticket can partly claim its ownership because of the critical role of its sparse structure in the final inference performance. %Breaking the sparse structure 
    \item We further propose a new mask embedding method that is capable of embedding ownership ``signatures" in the subnetwork's sparse structural connectivity (Fig.~\ref{fig:tesear}), without much affecting its performance. The signature is robust, e.g., it can be extracted and decoded even after pruning or fine-tuning attacks. Combined further with the trigger set-based method, our mask embedding method can work under both white-box and black-box verification frameworks. 
    \item We investigate several verification schemes, \textit{i.e.,} separate masks, embedding signatures, and embedding signatures with the trigger set. We show that these schemes are robust to the common removal and ambiguity attacks, as well as a new type of ``add-on" attacks. Extensive experiment results demonstrate their competence on protecting the winning tickets. For example, on ResNet-20, our verification framework can defend fine-tuning attacks intrinsically, as well as pruning attacks and as add-on attack under all levels of pruning ratios.

    %Our embedding signatures can defend against pruning attack under XX\% level, add-on attack under XX\% level. With trigger set enabled, our method can be further robustified to fine-tune attacks under XX\% level.
    
    %Numerically speaking, the detection rate is XX\% after fine-tuning attack and XX\% after pruning attacks. 
\end{itemize}

\vspace{-2mm}
\section{Related Work}
\vspace{-2mm}
\paragraph{Pruning and Lottery Ticket Hypothesis.} Pruning algorithms can be roughly classified into two types: pruning \textit{after} training and pruning \textit{before} training. Conventional pruning after algorithms assign scores to individual weights and remove weights with the lowest scores~\cite{tanaka2020pruning}. There are a large number of scoring algorithms of this kind, including weights magnitude~\cite{janowsky1989pruning,han2015learning}, Taylor coefficients~\cite{mozer1989skeletonization,lecun1990optimal,hassibi1993second,molchanov2016pruning}, and other variants~\cite{dong2017learning,yu2018nisp}. Pruning-before-training methods also play important roles in the community~\cite{frankle2018lottery}. However, it requires multiple cycles of training and pruning process~\cite{tanaka2020pruning}, making it sometimes tedious in practical use. Pruning-at-initialization methods alleviate such cost by pruning initial weights such as observing initial gradients of weights~\cite{lee2018snip,wang2019picking}, but the quality of sparse networks found by these methods are mediocre in return. 

A representative pruning-before-training method is the lottery ticket hypothesis~\cite{frankle2018lottery}, which hypothesizes the existence of a sparse network in dense networks that can be trained to match the test accuracy of dense networks in isolation. It was later verified that the original hypothesis was too strong, and early rewind techniques~\cite{frankle2020linear} help scale up the original version. Later on, the lottery ticket hypothesis and its variants have been explored and extended to various fields~\cite{gale2019state,pmlr-v139-zhang21c,chen2020lottery,chen2020lottery2,yu2019playing,chen2021gans,ma2021good,gan2021playing,chen2021unified} such as image generation~\cite{kalibhat2021winning,chen2021gans,chen2021ultra} and natural language processing~\cite{gale2019state,chen2020lottery}. However, it is currently non-trivial to find winning tickets, especially at high sparsity since multiple train-prune-train processes are required, which suggests the practical value of protecting the found sparse networks. 

%\paragraph{Graph Measures and Metrics. } A wide range of graph measures and metrics have been studied in previous literature~\cite{wills2020metrics}. These metrics and measures are useful for comparison of graph structures, which has various applications in fields such as neuroscience~\cite{bassett2017network,de2014graph,fornito2016fundamentals}, cyber-security~\cite{chen2012smart,pasqualetti2013attack}, social network analysis~\cite{myers2014information} and bioinformatics~\cite{garroway2008applications}. There are mainly three types of metrics~\cite{wills2020metrics}: spectral metric, matrix metric, and feature-based metric. The feature-based metrics calculate specific “features” of the graph, such as the degree distribution, betweenness centrality distribution, diameter, number of triangles, number of $k$-cliques, etc. They can characterize key properties of a graph from different aspects, which we will apply in our work to establish the connection between sparse masks and graphs and discover their properties. 
\vspace{-2mm}
\paragraph{Ownership Verification.} Ownership verification has drawn attention from both the industry and academia. Many works have been proposed to address IP protection. One most popular way is to use watermarking algorithm: \cite{uchida2017embedding} proposed to embed watermarks in the form of bits into deep networks' weights by an additional regularization term. \cite{darvish2019deepsigns} embedded information into model weights by regularizing on the signs of weights. Besides watermarking on weights, \cite{adi2018turning,le2020adversarial} embedded watermarks in the labels of certain examples in a \textit{trigger set}, which makes it possible to extract watermarks through a service interface without directly accessing the models' weights (black-box setting). Following somehow different pathways, \cite{fan2019rethinking} proposed a passport-based approach that encodes signatures with special passport layers. \cite{zhang2020passport} presented passport-embedded normalization whose parameters are associated with signatures. However, none of these methods leverages structural information for ownership verification besides assuming general dense networks. For sparse networks, the sparse mask patterns represent the key information and can vary across models and tasks. 

%are always different across different models and tasks. To fill the literature gap, we explore such direction in this paper. 

\vspace{-2mm}
\section{The Lottery Ticket Claims its Ownership}
\vspace{-1mm}
%\subsection{Extremely Sparse Winning Tickets}
\subsection{The Lottery Ticket Hypothesis}
\vspace{-1mm}
\ding{182} \textbf{Sparse Masks, Subnetworks and Winning Tickets.} We define a neural network parameterized by $\mathbf{W}$ as $\mathbb{N}[\mathbf{W}](\cdot)$. With a slight abuse of notion, we define a subnetwork of $\mathbb{N}[\mathbf{W}]$ as $\mathbb{N}[\mathbf{W}, \mathbf{M}]:=\mathbb{N}[\mathbf{W}\odot\mathbf{M}](\cdot)$ where $\mathbf{M}$ is a sparse mask whose shape is the same as $\mathbf{W}$ but the value of each entry in which can only be either 0 or 1. Given $\mathbf{W}_0$ the initialization of $\mathbb{N}$, if $\mathbb{N}[\mathbf{W}_0, \mathbf{M}]$ can be \textbf{re-trained} to \textit{match} the test performance of the dense model training from $\mathbb{N}[\mathbf{W}_0]$, we call $\mathbb{N}[\mathbf{W}_0, \mathbf{M}]$ a \textbf{winning ticket}. The term re-train above is used to distinguish between the training process to find the winning ticket. The criterion for matching can be set as, e.g., no lower than $1\%$ than dense models' performance. %After deriving the winning ticket, one need to \textbf{re-train} the winning ticket to  
\ding{183} \textbf{Sparsity Comparison.} The sparsity of a sparse mask $\mathbf{M}$ can be defined as $\textrm{spar}(\mathbf{M}):=\|\mathbf{M}\|_0 / \|\mathbf{M} + 1\|_0$ where $\|\cdot\|_0$ represents the non-zero values of the input matrix, and the relative sparsity can be defined as $\textrm{rspar}(\mathbf{M}_1, \mathbf{M}_2):=\textrm{spar}(\mathbf{M}_1)/(\textrm{spar}(\mathbf{M}_2) + \epsilon)$. We call a sparse mask $\mathbf{M}_1$ is \textit{sparser} than $\mathbf{M}$ if and only if $\|\mathbf{M}_1\|_0 < \|\mathbf{M}\|_0$. A \textit{sub-mask} $\mathbf{M}_2$ is a mask that has the same shape as $\mathbf{M}$ satisfying that all the elements in $\mathbf{M} - \mathbf{M}_2$ are non-negative. 
\ding{184} \textbf{Extremely Sparse Condition.} Given a sparsity difference threshold $t$, we call $\mathbb{N}[\mathbf{W}_0, \mathbf{M}_e]$ an \textbf{extremely sparse winning ticket} (or referred to as an \textbf{extreme ticket} hereinafter for conciseness) if $\mathbb{N}[\mathbf{W}_0, \mathbf{M}_e]$ can match the performance of $\mathbb{N}[\mathbf{W}_0]$, but pruning the model (\textit{i.e.}, increase the sparsity of $\mathbf{M}_e$) $t \times 100\%$ further cannot produce a winning ticket. In our experiments, we set $t$ to be $0.01$. 
\vspace{-1mm}
\subsection{Verification Framework for Extremely Sparse Winning Tickets}
\vspace{-1mm}
%In this section we will formulate the ownership verification framework of winning tickets. 
%We introduce two protection scenarios: protecting the winning ticket before training $\mathbf{M}\odot\mathbf{W}_0$, or protecting the trained winning ticket $(\mathbf{M}\odot\mathbf{W})_\mathrm{T}$. 
%\paragraph{Definition.}
A ownership verification framework for extremely sparse winning tickets can be formulated as a tuple $\mathcal{V} = (ME,WE,F,V,I)$, each item of which is a process:
\vspace{-2mm}
\begin{itemize}[leftmargin=*]
     \item A \textit{mask embedding} process $ME(\mathbf{M}_0, \mathbf{s})$ (optionally) for sparse masks. $\mathbf{s}$ is an optional string that can be encoded into masks. The output of this process is a new mask $\mathbf{M}$. $\mathbf{M}$ can either be a mask with $\mathbf{s}$ embedded or contains other structural information that is useful for ownership verification. We call the verification method with $ME(\mathbf{M}_0, \mathbf{s})$ enabled a ``mask-based'' method.
    \item A \textit{weight embedding} process $WE$($\mathbf{D}_\mathrm{tr}$, $\mathbf{T}$, $\mathbf{s}$, $\mathbb{N}[\cdot]$,$L$, $\mathbf{W}_0$, $\mathbf{M}$), which is a learning process for the lottery ticket model. $\mathbf{D}_\mathrm{tr}=\{\mathbf{x}, y\}$ is the training dataset, $\mathbf{T}=\{\mathbf{T}_x, \mathbf{T}_y \}$ is an optional trigger set provided to the training process, $\mathbf{s}$ is an optional signature for the weights embedding process. $L$ is the loss function for model training (usually the cross-entropy loss), $N[\cdot]$ defines the model structure, $\mathbf{W}_0$ is the initialization of weights, and $\mathbf{M}$ is the sparse mask for the model. The output of $E$ is a model $\mathbb{N}$ with sparse weights $\mathbf{W} \odot \mathbf{M}$ where $\mathbf{W}$ represents the trained weights. The trigger set $\mathbf{T}$ (or/and) the signature $\mathbf{s}$ are embedded in $\mathbf{W} \odot \mathbf{M}$ after this process and can be verified with the verification process introduced next.   %Under the setting of lottery ticket hypothesis, we also need to record the initialization weights, which are denoted by $\mathbf{W}_0$. 

    %\item A \textit{watermarking} process is an optional procedure to generate information for ownership verification \textit{after training}. %We exemplify two methods: a key-based method and a graph-based method. For the key-based method, the process is to separate the mask $\mathbf{M}$ into two disjoint sub-masks, one of which contains much more active weights than another.  generate a new mask $\mathbf{K}\in\{0,1\}^{d}$ such that $\sum_{i=1}^d \mathrm{K}_i > 0$ and $\mathrm{M}_i-\mathrm{K}_i \ge 0,i=1,2,3,\dots,d$. This guarantees that $\mathbf{K}$ is a \textit{sub-mask} of $\mathbf{M}$. We call $\mathbf{K}$ a $\textit{key}$, and denote the difference $\mathbf{M}-\mathbf{K}$ by $\mathbf{M}'$, which is also a valid mask since $\mathrm{M}_i-\mathrm{K}_i \ge 0$.
    
    %For the graph-based method, the process is to calculate values/statistics that are associated with the graph structures. 
    
    \item A \textit{fidelity evaluation} process $F(\mathbb{N}[\cdot], \mathbf{W}, \mathbf{M}, \mathbf{D}_\mathrm{te},\mathcal{A}_f, \epsilon_f)$ is to evaluate whether the performance  discrepancy of model $\mathbb{N}[\cdot]$ is less than a pre-defined threshold $\epsilon_f$, \textit{i.e.,} $|\mathcal{A}(\mathbb{N}[\mathbf{W},\mathbf{M}], \mathbf{D}_\mathrm{tr})-\mathcal{A}_f|<\epsilon_f$, in which $\mathcal{A}(\cdot, \cdot)$ is the inference performance on the test dataset $\mathbf{D}_\mathrm{te}$, and $\mathcal{A}_f$ is a target inference performance associated with the model. 
    \item A \textit{verification} process $V(\mathbb{N}[\cdot],\mathbf{W}, \mathbf{M},\mathbf{T},\mathbf{s},\epsilon_s)$ checks whether the sparse mask $\mathbf{M}$ or the trigger set $\mathbf{T}$ can be successfully verified for a given model $\mathbb{N}[\cdot]$. For the mask-based methods, the process is to check if $\mathbf{M}$ and $\mathbf{s}$ matches by evaluating $N[\cdot, \mathbf{M}]$ on $\mathbf{D}_\mathrm{te}$ to see if the performance gap is smaller than a pre-defined threshold $\epsilon_s$, and(or) extract information in $\mathbf{M}$ and compare it with $\mathbf{s}$. For the trigger set-based methods, an inference is first executed on the trigger set images $\mathbf{T}_x$, and then the prediction will be compared with trigger set labels $\mathbf{T}_y$ to see if the false detection rate is lesser than a threshold $\epsilon_s$~\cite{fan2019rethinking}.    
    %For the key-based scheme, $V$ will judge if $\mathbf{K}$ equals $\mathbf{M}-\mathbf{M}'$ and if $\mathbf{K}\odot\mathbf{M}'=\mathbf{0}$. For the graph-based scheme, $V$ will verify the graph property calculated on $\mathbf{W}$; For the trigger-set based scheme, $V$ will perform an inference process on the trigger set $\mathbf{T}$, and \textcolor{red}{check whether the prediction corresponds to the designated labels with a false detection rate smaller than a threshold $\epsilon_s$.} 
    \item An \textit{invert} process $I(N[\mathbf{W}, \mathbf{M}], \mathbf{T}, \mathbf{s})$ exists and will enable a successful ambiguity attack~\cite{fan2019rethinking} if:
    a) a set of new trigger set $\mathbf{T}'$, a new signature $\mathbf{s}'$, or a new mask $\mathbf{M}'$ can be reverse-engineered for the given mask $\mathbf{M}$ and weights $\mathbf{W}$.  
    b) the forged $\mathbf{T}'$, $\mathbf{s}'$, $\mathbf{M}'$ can be verified with respect to $\mathbf{M}$ and $\mathbf{W}$. 
    c) the fidelity evaluation outcome $F(\mathrm{N}[\cdot],\mathbf{W},\mathbf{M}', \mathbf{D}_\mathrm{te},\mathcal{A}_t,\epsilon_f)$ remains True.
\end{itemize}
\vspace{-2mm}
The high-level definitions above are general and can work with any concrete implementation. We will introduce several methods that use sparse structural information in the next following sections. 

\vspace{-2mm}
\subsection{Structural Information As Signatures}
\vspace{-2mm}
Our motivation is originated from the nature of winning tickets that the sparse structure of winning tickets is critical to their performance. As the sparse masks found by IMP outperform other pruning methods by clear margins \cite{frankle2020pruning}, incorrect masks will lead to degraded test accuracy. 
%, which is different with the previous practice on ownership verification: 
%It is known that the sparse masks found by IMP outperform other pruning methods by clear margin~\cite{frankle2020pruning}, so forging or modifying the sparse masks will deteriorate the performance of sparse models. 
In the next few sections, we demonstrate how to use the sparse structure of extreme tickets, \textit{i.e.,} both the sparse masks and weights, to perform ownership verification under different verification schemes. The ownership verification can be performed in two different scenarios: (a) protecting the sparse masks of the extremely sparse winning tickets; and (b) protecting the trained extremely sparse winning tickets. 
\vspace{-0.5em}
\subsubsection{Protecting the Sparse Masks: Splitting Signature from Sparse Model}
Such sparse masks play a crucial role in achieved outstanding generalization~\cite{frankle2018lottery} and transferability~\cite{chen2020lottery,chen2020lottery2}, and thus draws our attention to prevent them from being illegal distributed or used.
% \textcolor{red}{[TODO] explain why protecting the sparsity mask. Aspects: transferability, generalizability. } %Therefore, we propose a mask-based method to protect such an important asset. 
Given a fixed initialization, correct masks are essential for training the extremely sparse winning tickets to match the performance of the dense network. If we split the sparse masks into two parts, neither part is intact and correct so neither can be trained to match the performance of the dense model with the given initialization. Recall the mechanism of one lock can be unlocked by one key generally, we adopt the concept of keys and locks and propose a new ownership verification method for the masks of extremely sparse winning tickets. 
%The severity of performance degradation will scale up with the number of incorrect entries in the set of masks and initial weights. Therefore, if we split the sparse masks into two parts, neither can be trained to match the performance of the dense model, which resembles the mechanism of key and lock where one ``lock'' can generally be unlocked by one ``key''. This mechanism inspires us to propose a new ownership verification method. 

Denote the sparse mask of extremely sparse winning ticket by $\{\mathbf{M}_l\}_{l=1}^N$ and the weights by $\{\mathbf{W}_l\}_{l=1}^N$, where $N$ is the number of layers. To sparsify a model, $\{\mathbf{M}_l\}_{l=1}^N$is applied to the model's weight $\{\mathbf{W}_l\}_{l=1}^N$ by conducting an element-wise product ($\{\mathbf{W}_l\odot \mathbf{M}_l\}_{l=1}^N$). Our goal is to find \textit{key masks} \textit{i.e.,} sub-masks $\{\mathbf{M}_l^s\}_{l=1}^N$ that contain as few elements as possible while the performance of the sparse network with the \textit{locked masks}, \textit{i.e.,} the remaining masks, degrade as much as possible. Meanwhile, fewer elements in key masks reduce the cost of storing, distributing, and using the key masks.
% , and the performance degradation will accentuate the need for correct key masks. 

%We name $\{\mathbf{M}_l^s\}_{l=1}^N$ \textit{key masks} and the remaining masks $\{\mathbf{M}_l - \mathbf{M}_l^s\}_{l=1}^N$ \textit{locked masks}. Training with locked masks cannot match the performance of the dense network since incorrect masks are used, so one must provide the key masks for appropriate training results, accentuating the need for key masks. 

We next describe the algorithms needed to discover key masks. An algorithm is used to split the masks of extremely sparse winning tickets ($\{\mathbf{M}_l\}_{l=1}^N$) into key masks $\{\mathbf{M}_l^s\}_{l=1}^N$ and locked masks under the constraint of $\mathrm{rspar}(\{\mathbf{M}_l^s\}_{l=1}^N, \{\mathbf{M}_l\}_{l=1}^N) < n_s$. $n_s$ is a hyper-parameter controlling the relative sparsity of the key masks. Score functions are used to decide which part should be split into the key masks. The pipeline is described in Algorithm~\ref{alg:alg1}.

\begin{algorithm}[h]
\LinesNumbered
\SetAlgoLined
\SetKwInOut{Input}{input}
\SetKwInOut{Output}{output}
\Input{A sets of masks $\mathbf{M}=\{\mathbf{M}_l\}_{l=1}^N$, initialization weights $\mathbf{W}=\{\mathbf{W}_l\}_{l=1}^N$, number of non-zero elements $n$, and a function \texttt{score}$(\cdot)$ for scoring. }
\Output{Key masks $\{\mathbf{M}_l^s\}_{l=1}^N$ and locked masks $\{\mathbf{M}_l - \mathbf{M}_l^s\}_{l=1}^N$}
Derive the score matrices by applying score$(\cdot)$ over $\{\mathbf{W}_l\odot\mathbf{M}_l\}_{l=1}^N$ and get $\{\mathbf{S}_l\}_{l=1}^N$

Set the values of entries in $\{\mathbf{S}_l\}_{l=1}^N$ to negative infinity if the corresponding entries at the same position in $\{\mathbf{W}_l\odot\mathbf{M}_l\}_{l=1}^N$ is zero (\textit{i.e.}, already pruned). 

Calculate the $n$\ts{th} largest number across the score matrix $\{\mathbf{S}_l\}_{l=1}^N$ and record it as $T$.

Set $\mathbf{M}_l^s\gets I_{\mathbf{M}_l > T}$ and let the key masks be $\{\mathbf{M}_l^s\}_{l=1}^N$. The comparison between $\mathbf{M}_l$ and $T$ $(\mathbf{M}_l > T)$ is performed element-wise. 

\caption{Splitting Key Masks}
\label{alg:alg1}
\end{algorithm}

We study several score functions in our experiments: 1) One-Shot Magnitude (OMP): the absolute values of each weight; 2) Edge-Weight-Product (EWP)~\cite{patil2020phew} which measures the importance of paths from models' input to output. The EWP score is defined as the multiplication of weights along the paths; 3) Edge betweenness centrality (Betweenness). The edge betweenness centrality measures the importance of each edge inside a graph. For convolutional layers, we define the weight of each ``edge'' to be the summation of absolute values of each element; and 4) random scoring. 

\subsubsection{Protecting the Trained Tickets: Embedding Signature into Sparsity Masks}

Another scenario is to protect the trained extremely sparse winning ticket since a superior performance on certain large-scale datasets usually comes with a huge economic and ecological cost.
% \textcolor{red}{[TODO] explain why protecting the trained ticket. Aspects: Large scale datasets, repeatedly training brings economic and ecological cost. }
Although directly splitting the masks provides a solution to the ownership verification problem, it has some drawbacks. It delivers extra cost to users since they need to recover the masks. Such a method is also intrusive and requires additional responsibility from the users' side for storing the key masks safely. To render the extreme tickets capable of self-verification and free of key masks
%To embed owner-related information in models' sparse structure
, we propose a novel pruning method that is able to ``absorb'' secret information (\textit{e.g.}, signatures) into models' sparse masks. The core concept is to enforce the sparsity masks to follow certain ``0-1'' patterns, which can be extracted from masks and further decoded back to the original form of information.

A function \texttt{encode}$(\cdot)$ is used to transform a string $\mathbf{s}$ into a matrix  $\mathbf{M}_s \in \{0, 1\}^{d_1 \times d_2}$ which we call \textit{signature mask}. Our goal is to embed $\texttt{encode}(\mathbf{s})$ into the sparsity masks $\{\mathbf{M}_l\}_{l=1}^N$. One critical question is where to embed the signature mask $\mathbf{M}_s$. Empirically, low-level convolutional layers are less sparser, which means they are more unlikely to be pruned. Therefore, information embedded in the low-level convolutional layers is more difficult to be removed if using the pruning method. Based on such observation, we decide to embed $\mathbf{M}_s$ in low-level convolutional layers. To minimize mask changes, we first find a region in $\{\mathbf{M}_l\}$ with the highest similarity with $\mathbf{M}_s$ and tune the sub-mask of that region. For masks that have a dimension of two, we directly replace the region with $\mathbf{M}_s$% and then make an adjustment to hold the layer-wise sparsity; for masks that have a dimension of more than two, we raise their dimension by using random connections. Our detailed workflow is shown in Algorithm~\ref{alg:alg2}.
; for masks that have a dimension of more than two, we raise their dimension by using random connections. Our detailed workflow is shown in Algorithm~\ref{alg:alg2}. 

The choices of function \texttt{encode}$(\cdot)$ are various but there is one common choice in our daily life: QR code~\cite{soon2008qr}. QR code has multiple advantages: 1) QR code is naturally seen as a pattern with only zeros and ones; 2) QR code has the ability to correct the error if the code is dirty or damaged. For example, the H correction level can tolerate up to 30\% of error~\cite{soon2008qr}; 3) QR codes can be small in size which can be easily fit into sparse masks. The size of the QR code generated can be as small as $21\times21$ while the numbers of channels in convolutional kernels in deep learning models are typically greater than $21$, showing an abundant space for fitting the QR code in inside models' sparsity masks; and 4) The QR code \textbf{without} the finder, alignment and version patterns are imperceptible when fitted into sparse masks since there are no ``regular'' patterns left. Based on these merits, we choose \texttt{encode}$(\cdot)$ to be the QR code generation function. Specifically, the \texttt{encode} function we use will return a QR code without finder, alignment, and version patterns. When extracting the code, the above patterns will be added back for decoding the credential information behind the QR code.

\begin{algorithm}[h]
\LinesNumbered
\SetAlgoLined
\SetKwInOut{Input}{input}
\SetKwInOut{Output}{output}
\Input{A set of masks $\mathbf{M}=\{\mathbf{M}_l\}_{l=1}^N$, a signature $\mathbf{s}$}
\Output{A set of masks with signature embedded $\{\mathbf{M}_l^e\}_{l=1}^N$}
Calculate $\mathbf{M}_s\gets \texttt{encode}(\mathbf{s})$.

Squeeze each $\mathbf{M}_l$ into a two-dimensional matrix $\mathbf{M}_l^f$ by setting $(\mathbf{M}_l^f)_{ij}  = \mathbb{I}_{\|(\mathbf{M}_l)_{ij}\|_0 > 0 }$.

Calculate the similarity (percentage of matched 0-1 patterns) between each $\mathbf{M}_l^f$ and $\mathbf{M}_s$ and name the one with the largest similarity $\mathbf{M}_\textit{max}^f$.

%locate the position having the most similarity in the high-level convolutional layers and 

Change the dimension of $\mathbf{M}_s$ and fit it into $\mathbf{M}_\textit{max}^f$ to the region where the similarity is the largest. 

\caption{Embed Signature Into Sparse Masks}
\label{alg:alg2}
\end{algorithm}

%\begin{figure}[h]
%    \centering
%    \subfloat[Embed code in the first layer of ResNet-18]{\includegraphics[width=0.4\linewidth]{Figs/vis.png}}
%    \subfloat[Embed in the third layer of ResNet-20]{\includegraphics[width=0.4\linewidth]{Figs/res20s_cifar10_vis.pdf}}
%    \caption{Embedding QR code in ResNet-18 and ResNet-20. The information hidden in QR code is the string ``\textit{Th1sIs4siGNaTure}''. }
%    \label{fig:QRcode}
%\end{figure}

\subsection{Ownership Verification with Sparse Structural Information}
\vspace{-2mm}
Next, we propose three different verification schemes based on sparse structural information, as summarized in Table~\ref{tab:my_label}. Under our unique context, we further introduce a new \textit{Add-on Attacks} which aims to create ambiguity against lottery verification by ``recovering" several pruned weights and manipulating the sparsity patterns. More details can be found in Appendix~\ref{sec:more_methods}. 

%By utilizing our graph-based approach, we design four ownership verification schemes which are listed in the next few paragraphs. 

%Two different scenarios are studied: mask ownership verification and trained weight ownership verification. 
\vspace{-2mm}
\paragraph{Scheme $\mathcal{V}_1$: Distribute the extreme tickets with key masks.} Scheme $\mathcal{V}_1$ is designed to protect the sparsity masks. We separate the sparsity mask $\mathbf{M}$ into two parts: $\mathbf{M}_\mathrm{l}$ and $\mathbf{M}_\mathrm{s}$, where $\mathbf{l}$/$\mathbf{s}$ subscripts denote ``large''/``small'', respectively. The small mask is sparser than the large one, which is used as the key mask while the large counterpart is the locked mask.
We apply these two masks on weights and get two separate parts $(\mathbf{W}\odot\mathbf{M}_\mathrm{l}, \mathbf{W}\odot\mathbf{M}_\mathrm{s})$.
%The two corresponding weights, $\mathbf{W}\odot\mathbf{M}_\mathrm{l}$ and  $\mathbf{W}\odot\mathbf{M}_\mathrm{s}$ are distributed to users. 
Before re-training, legitimate users should merge the two weights by adding them up to recover the original sparse weights $\mathbf{W}\odot\mathbf{M}$. The ownership can be automatically verified by the inference performance since an incorrect provided mask-weight pair will deteriorate accuracies after re-training.%regarding the inference performance. 

%\paragraph{Scheme 2: Graph measures as signatures. }
%Scheme 2 uses pre-calculated graph measures before distributing the model as verification certificates. The model with initialization or the trained weights is distributed to legitimate users. The users does not need any further process before using the model to infer or fine-tune since the complete weights are given. The ownership verification process will be carried out only upon requested by law enforcement department. The process requires direct access to the weights: it calculates the graph metrics layers by layers and compares them with the pre-calculated graph measures.  

%Compare with Scheme 1, this scheme is easier to use since there is no additional training cost and no extra processing before inference or fine-tuning. In the meantime, the ownership verification for this scheme is robust to removal attacks (explain). However, the disadvantages of this scheme is clear, that is one cannot choose the signature to embed in the model since the signature is exclusively derived from the sparsity masks. To overcome such a drawback, we propose a different pruning method which is capable of embedding any signature to the graph while not harming the performance of sparse models. 
\vspace{-2mm}
\paragraph{Scheme $\mathcal{V}_2$: Embed signatures in sparse masks.}
%Using vanilla IMP methods to find winning tickets at extreme sparsity does not allow for embedding additional information such as signatures into the sparsity masks. 
We apply the signature mask embedding method to embed credentials into the extreme ticket. Then we train the model and dispatch it to legitimate users. No further action is required at the users' side. 
%by introducing a novel pruning method, the sparse structure now can contain more information using sparsity masks. No pre-processing process is required for inference and fine-tuning. 
For the verification process, one can use extract the signature from the sparse model and validate the ownership of the extreme tickets. Compared with Scheme $\mathcal{V}_1$, Scheme $\mathcal{V}_2$ is more user-friendly since no extra action is performed at the users' side. The application scenarios of Scheme $\mathcal{V}_1$ and $\mathcal{V}_2$ are also different: the latter focuses on protecting the trained weights. It also shows great defense ability towards removal and ambiguity attacks. However, this scheme works under the white-box verification setting only, which means that access to models' weights has to be assumed. To overcome that assumption, we combine Scheme $\mathcal{V}_2$ with a trigger set-based method and propose Scheme $\mathcal{V}_3$ in the next section. 
\vspace{-2mm}
\paragraph{Scheme $\mathcal{V}_3$: Combining trigger set-based methods.}
Scheme $\mathcal{V}_3$ is more sophisticated than Scheme $\mathcal{V}_2$ as a set of trigger images and labels are used during the (re-)training process. With the help of this trigger set, Scheme $\mathcal{V}_3$ is now capable of black-box verification. By using remote calls of service APIs, the owner can first probe and claim the ownership in a black-box regime and further request a white-box verification if the black-box mode has raised a red flag. The white-box verification part for Scheme $\mathcal{V}_3$ remains the same as Scheme $\mathcal{V}_2$. 
\vspace{-2mm}
\section{Experiments}
%\vspace{-1em}
In this section, we will list the details of our experiments and show the results to prove the effectiveness of our ownership verification methods, as well as the robustness to removal attacks ($e.g.$, model pruning, fine-tuning, and add-on attacks) and ambiguity attacks ($e.g.$, fake paths, fake code). 
\vspace{-1em}
\paragraph{General Settings.}~\label{sec:setting}
We use three networks architectures (ResNet-20, ResNet-18 and ResNet-50) and two benchmarks (CIFAR-10~\cite{krizhevsky2009learning} and CIFAR-100~\cite{krizhevsky2009learning}) in our experiments. For all experiments, we follow the same (re-)training and testing protocol. The optimizer we use is an SGD optimizer with a momentum factor of 0.9 and a weight decay factor of 1e-4 for ResNet-20 and ResNet-18, and 5e-4 for ResNet-50. We train the model for 182 epochs with an initial learning rate of 0.1, and we decay the learning rate at 91\ts{st} and 136\ts{th} epoch by 0.1. We use a late rewinding technique that rewinds the weight to the 3\ts{rd} checkpoint. %A detailed setup of our experiments is deferred to the supplementary files. 
More results on ResNet-50 are deferred to Appendix~\ref{sec:more_results}. Our experiments are run with 16 pieces of NVIDIA RTX GeForce 2080 Ti. 

\vspace{-1em}
\paragraph{Types of Attacks and Trigger Set.}
We study two types of attacks: 1) removal attacks. This type of attack aims at removing the embedded watermarks from the model's weights or data. Available methods for removal attacks include pruning, which removes a proportion of parameters of the model, and fine-tuning, which performs training on new data for a few steps. Both methods can modify the weights and potentially make the watermark undetectable. 2) ambiguity attacks. This type of attack aims at confusing the verification schemes, \textit{i.e.,} no one can tell which is the real watermark/signatures. This type of attack needs techniques like reverse engineering and does not have a certain form. We explore several attack methods in our paper. 

The trigger set we use is the same as the trigger set used in \cite{adi2018turning}, which contains abstract images that are different than the training images. 

\vspace{-1em}
\paragraph{New type of attack: Add-on Attacks.}
Although the extremely sparse winning tickets are naturally robust to fine-tuning and pruning attacks due to their unique properties, the verification schemes based on the sparse structure will potentially suffer from another kind of attacks, i.e., trying to ``recover'' some pruned weights and change the sparse structure of the extremely sparse winning tickets. We name it \textit{add-on attacks}. Such a new attack type targets mask-based verification schemes and aims at creating ambiguity against verification. 
%Such a attack will not necessarily impair the performance but might cause ambiguity when performing ownership verification if the verification method is based on the sparse structure. 

We propose a pipeline defending against such attacks. We can first prune weights whose magnitudes are smaller than $t$. $t$ is known to the owner of the model since the owner has the authentic sparse masks. This can detect any noise with magnitude smaller than $t$, that the attackers add to the mask.  For noises of moderate level, their magnitudes become comparable to with the benign weights, hence the prediction quality will be significantly degraded as the noise increases.

%In this section we report the experiment results on winning tickets where the inference performances of various schemes are compared in terms of robustness to both removal attacks and ambiguity attacks. %We study four different rewinding settings for fully study the results. 
\vspace{-1mm}
\subsection{Finding Extreme Tickets}
\vspace{-2mm}
%\begin{table}[htbp]
\begin{wraptable}{r}{7cm}
    \centering
    \vspace{-4mm}
    \caption{\small Performance of dense models and extremely sparse winning tickets, and the pruning specification. %We report the results of two models (ResNet-20, ResNet-18) on two benchmarks (CIFAR-10, CIFAR-100). 
    The performance are expressed in terms of the test accuracy of the dense model and the extremely sparse winning ticket. The pruning specification includes the proportion of remaining weight as well as the number of pruning with four different pruning ratios (in brackets). }
    \vspace{-2mm}
    \resizebox{1\linewidth}{!}{
    \begin{tabular}{c|c|c|c}
        \toprule
        Model & Dataset &
        Performance & Pruning Specification \\ 
        \midrule
        ResNet-20 & CIFAR-10 & 91.67\%,91.66\% & 19.369\% (5,1,8,1)\\
        ResNet-20 & CIFAR-100 & 66.36\%,66.39\% & 19.901\% (6,0,4,7)\\
        \midrule
        ResNet-18 & CIFAR-10 & 93.67\%,93.60\% & 1.236\% (18,3,0,6) \\
        ResNet-18 & CIFAR-100 & 72.44\%,72.59\% & 2.251\% (17,0,0,0) \\
        %\midrule
        %ResNet-50 & CIFAR-10 & & & \\
        %ResNet-50 & CIFAR-100 & & & \\
         \toprule
    \end{tabular}}
    \label{tab:extreme_winning_tickets}
%\end{table}
\vspace{-6mm}
\end{wraptable}
To find extreme tickets, multiple rounds of the train-prune-retrain process are usually required. Once the test performance of the currently trained model cannot match the performance of the dense model, we revert the pruning process back for one time and reduce the pruning ratio. The choices of pruning ratio of weights are $[0.2, 0.1, 0.05, 0.1]$. The results are shown in Table~\ref{tab:extreme_winning_tickets}. We also report the pruning specification, which includes the remaining weights of the extreme tickets, as well as the pruning times for each pruning rate we choose. 

Notice that ResNet-20 needs at least $15$ times of pruning before we identify the extreme tickets, and for ResNet-18, such number increases to $17$. Numerous pruning rounds exemplify the effort to find the extreme tickets and emphasizes the importance of protecting them.

\vspace{-2mm}
\subsection{Effectiveness of Different Schemes}
\begin{figure}[h]
    \centering
    \vspace{-2mm}
    \includegraphics[width=0.9\linewidth]{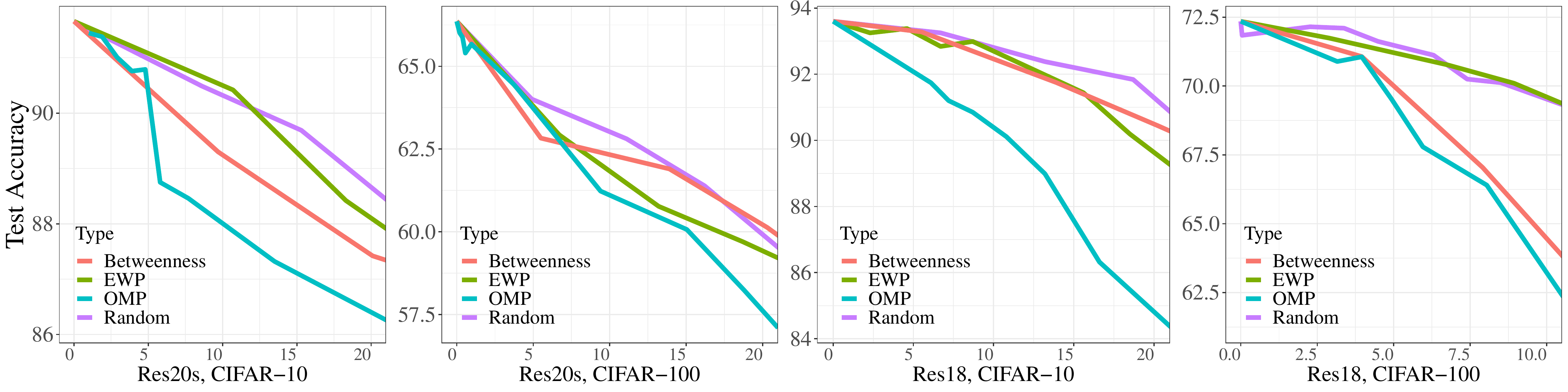}
    \vspace{-2mm}
    \caption{\small Effectiveness of Scheme $\mathcal{V}_1$: Re-training without key masks generated by four methods: Betweenness, EWP, OMP, Random (Paths). The $x$-axis is the relative sparsity w.r.t the extreme ticket.}
    \label{fig:scheme1}
    \vspace{-4mm}
\end{figure}

\vspace{-2mm}
\paragraph{Scheme $\mathcal{V}_1$}
To verify that extreme tickets without correct key masks will have degraded performance after retraining, we conduct experiments that directly re-train the models without the key masks on ResNet-20 and ResNet-18. Fig.~\ref{fig:scheme1} show the performance of extreme tickets after retraining. %The number of non-zero element in the key mask balance the trade-off between efficiency and the strength of protection: fewer ``1''s in key masks impose less stress on distribution and ``unlocking'' the locked masks while more ``1''s in key masks can increase the performance divergence. 
It can be seen that more ``1''s in key masks can increase the performance divergence. Different splitting functions do not make an essential difference, showing that the choice of \texttt{score} functions is flexible. %It can also be seen that an appropriate number of paths 

%\begin{figure}[h]
%    \centering
%    \includegraphics[width=1.0\linewidth]{Figs/pic2.pdf}
%    \caption{Effectiveness of Scheme 1: Trained }
%    \label{fig:extreme_sparsity}
%\end{figure}

\paragraph{Scheme $\mathcal{V}_2$}
To show that our proposal can embed information into models' sparse structures without significantly harming their performance, we conduct experiments to compare the performance before and after a signature string is embedded. Table~\ref{tab:effectiveness_scheme_2} shows the test accuracy of two different models. We can see from the table that the performance of extreme tickets with the string embedded is only slightly lower than the original one, which proves that using Scheme $\mathcal{V}_2$ endows owners the ability to embed information at a little cost of performance. %The information embedded in the sparse masks can be extracted easily. We show the QR code extracted from the graph in Figure~\ref{fig:extract_qr_code}. All the codes can be parsed into a string. 
\begin{table}[htbp]
\begin{minipage}[t]{0.47\linewidth}
    \centering
    \vspace{-4mm}
    \caption{\small Effectiveness of Scheme $\mathcal{V}_2$: performance of extreme tickets after embedding QR codes. We study two different models and compare their inference performance. The performance after embedding and the performance drop are reported (in brackets).}
    \resizebox{0.98\linewidth}{!}{
    \begin{tabular}{c|c|c}
    \toprule
    \multirow{2}{*}{Model} & \multicolumn{2}{c}{Accuracy after embedding}  \\
    \cmidrule{2-3}
     & CIFAR-10 & CIFAR-100 \\
    \midrule
    ResNet-20 & 91.37\% ($\downarrow$ 0.29\%) & 66.14\% ($\downarrow$ 0.25\%) \\
    ResNet-18 & 93.56\% ($\downarrow$ 0.11\%)& 72.35\% ($\downarrow$ 0.24\%)\\
    %ResNet-50 & & & \\
    \bottomrule
    \end{tabular}}
    \vspace{-6mm}
    \label{tab:effectiveness_scheme_2}
    \end{minipage}
    \hfill%
    \begin{minipage}[t]{0.47\linewidth}
        \centering
        \vspace{-4mm}
        \caption{\small Effectiveness of Scheme $\mathcal{V}_3$: ResNet-20 on CIFAR-10 and CIFAR-100. ESWT is the abbreviation of \textbf{E}xtremely \textbf{S}parse \textbf{W}inning \textbf{T}ickets.
        We re-train the two extremely sparse winning tickets with QR code embedded found with the trigger set enabled. }
        \resizebox{0.98\linewidth}{!}{
        \begin{tabular}{c|c|c}
        \toprule
        \multirow{2}{*}{Model} & \multicolumn{2}{c}{Test Accuracy} \\
        \cmidrule{2-3}
         & CIFAR-10 & CIFAR-100 \\
        \toprule
        ESWT & 91.66\% (16.0\%) & 66.36\% (0.0\%) \\
        %ESWT + \textbf{T} & 91.49\% (93.0\%) & 65.81\% (89.0\%) \\
        ESWT + $\mathbf{M}_s$ + \textbf{T} & 91.46\% (91.0\%) & 65.39\% (90.0\%) \\
        \bottomrule
        \end{tabular}}
        \vspace{-6mm}
        \label{tab:trigger}
    \end{minipage}
\end{table}

%\begin{figure}[h]
%    \centering
%    \subfloat[ResNet-20]{\includegraphics[width=0.3\linewidth]{Figs/res20s_cifar10_vis.pdf}}
%    \subfloat[ResNet-20]{\includegraphics[width=0.3\linewidth]{Figs/res20s_cifar10_vis.pdf}}
%    \subfloat[ResNet-20]{\includegraphics[width=0.3\linewidth]{Figs/res20s_cifar10_vis.pdf}}
%    \caption{Visualization of the masks where QR codes are embedded. }
%    \label{fig:extract_qr_code}
%\end{figure}

\paragraph{Scheme $\mathcal{V}_3$}
We use a set of trigger images during the re-training of extremely sparse winning tickets under Scheme $\mathcal{V}_3$. The inference performance on the original task (\textit{i.e.}, CIFAR-10 or CIFAR-100) should not be greatly affected with trigger sets enabled. Table~\ref{tab:trigger} shows the inference performance on both the original images and the trigger set. We can see that the performance only drops 0.2\% on CIFAR-10 and 0.97\% on CIFAR-100 for ResNet-20, while the detection rates of the trigger set images are high (91.0\% on CIFAR-10 and 90.0\% on CIFAR-100). At the same time, the extreme tickets trained from CIFAR-10 and CIFAR-100 without a trigger set can only have trigger set accuracies of 16.0\% and 0.0\%, respectively. This suggests the Scheme $\mathcal{V}_3$ can work as expected, \textit{i.e.,} perform well on the trigger set while not significantly harming the performance on the original dataset.  

% \begin{table}[htbp]
    
% \end{table}

\vspace{-2mm}
\subsection{Robustness Against Removal Attacks}
\vspace{-1mm}
%In this section we will attack Scheme 2 and Scheme 3 using removal attacks. Scheme 1 are removal attacks-free. 

\paragraph{Fine-tuning Attacks}

%Table~\ref{tab:finetune} shows that the distribution of graph measures are matched nearly 100\% for all ownership verification schemes in the original task. Even after fine-tuning the winning tickets, the graph measures still remain a similar distribution. We use XXXX (Maybe KL Divergence?) to compare the two distributions of graph measures, and we claim a match only if the xxxx is smaller than xxxxx. 

Fine-tuning the model can only change the values of weights while not changing the sparse structure of extreme tickets. As a consequence, Scheme $\mathcal{V}_2$ and $\mathcal{V}_3$ is resistant to fine-tuning attacks under the white-box verification setting. 

For Scheme $\mathcal{V}_1$, users are required to provide the key masks to recover the correct masks and then re-train the extreme tickets. One key property we need to verify is that attackers cannot bypass the requirement of key masks by fine-tuning the model on a new dataset. To this end, we conduct transfer experiments described as follows: on CIFAR-100, we train the model with the locked mask generated on the extreme tickets identified on CIFAR-10; on CIFAR-10, we conduct a similar experiment with the locked mask from CIFAR-100. %train the extremely sparse winning tickets found on CIFAR-10 to CIFAR-100, and the same experiment on a reversed direction (. 
The results are shown in Table~\ref{tab:finetune}. From the table, we can see that even transferring the sparse mask cannot bypass the requirement of key masks. The performance gaps between the transferred model and the extreme tickets found on each set are greater than 3\% on both datasets, much higher than the 1\% criterion we set for matching performance. Such big gaps prove that the model after fine-tuning attacks is not useful in practice. 

\begin{table}[htbp]
\begin{minipage}[t]{0.48\linewidth}
    \centering
    \vspace{-3mm}
    \caption{\small Fine-tuning Attacks on Scheme $\mathcal{V}_1$: Transferring extreme tickets of ResNet-20 found on CIFAR-10/100. 10$\rightarrow$100 means transferring from CIFAR-10 to CIFAR-100 and vice versa. The percentage inside brackets denotes the relative sparsity of the key mask w.r.t the extremely sparse winning ticket. }
    \resizebox{0.90\linewidth}{!}{
    \begin{tabular}{c|c|c}
    \toprule
    \multirow{2}{*}{Model} & \multicolumn{2}{c}{Test Accuracy} \\
    \cmidrule{2-3}
    & 10$\rightarrow$100 & 100$\rightarrow$10 \\
    \toprule
    % \\
    OMP (5\%) & 59.80\% & 87.66\% \\
    EWP (5\%) & 60.27\% & 88.21\% \\
    Betweenness (5\%) & 59.61\% & 87.22\% \\
    \bottomrule
    \end{tabular}}
    \vspace{-1mm}
    \label{tab:finetune}
\end{minipage}
\hfill%
\begin{minipage}[t]{0.48\linewidth}
    \centering
    \vspace{-3mm}
    \caption{\small Pruning Attacks on Scheme $\mathcal{V}_3$: Performance of ResNet-20 on CIFAR-10/100 after pruning with different pruning ratios. The accuracy on CIFAR-10/100 are shown outside the brackets and the accuracy on trigger images are inside the brackets. }
    \resizebox{0.98\linewidth}{!}{
    \begin{tabular}{c|c|c}
    \toprule
    \multirow{2}{*}{Model} & \multicolumn{2}{c}{Accuracy} \\
    \cmidrule{2-3}
     & CIFAR-10 & CIFAR-100 \\
    \toprule
    %Dense model & & \\
    Original model & 91.46\% (91.0\%) & 65.39\% (90.0\%) \\
    Pruning 5\%  & 91.33\% (89.0\%)   & 64.78\% (91.0\%) \\
    Pruning 10\% & 90.66\% (90.0\%)   & 62.96\% (73.0\%) \\
    Pruning 20\% & 87.86\% (81.0\%)   & 50.14\% (16.0\%) \\
    Pruning 50\% & 33.04\% (18.0\%)   & 8.56\% (0.00\%) \\
    \bottomrule
    \end{tabular}}
    \vspace{-1mm}
    \label{tab:pruning_scheme_3}
\end{minipage}

\end{table}

We also have conducted experiments to study if Scheme $\mathcal{V}_3$ can resist fine-tuning attacks under black-box verification. We first retrain the extreme tickets under Scheme $\mathcal{V}_3$ on CIFAR-10/-100, and continue to fine-tune it on CIFAR-100/-10. The extreme tickets trained on CIFAR-10 can only achieve 61.59\% test accuracy on CIFAR-100, and the extreme tickets on CIFAR-100 can only achieve 88.21\% test accuracy on CIFAR-10. The strong bond between sparse structure (masks of extreme tickets) and datasets on which the extreme tickets we found brings performance drop when fine-tuning them on a new dataset, which devalues such attack and also highlight the robustness of the Scheme $\mathcal{V}_3$ against fine-tuning attacks. 

%Fine-tuning the model cannot change the sparse structures, so under the white 

\paragraph{Model pruning}

Pruning the model under Scheme $\mathcal{V}_1$ is meaningless since pruning cannot recover the full masks. So we focus on Scheme $\mathcal{V}_2$ and $\mathcal{V}_3$ for model pruning attacks. %One method to remove the QR code from the sparse masks is to prune the model. 
Pruning the trained model leaves more ``0'' in the trained model, which might change the extracted QR code and makes it unable to decode. To study if our model can resist the pruning attack, we conduct experiments with different pruning methods (one-shot magnitude and random pruning) and different pruning ratios (5\%, 10\%, 20\%, 30\%, 50\%). 

We first examine our proposal for black-box verification (Scheme $\mathcal{V}_3$). In Table~\ref{tab:pruning_scheme_3} we show the results of Scheme $\mathcal{V}_3$ against pruning attacks. The accuracy on trigger set images drops after the accuracy on the original dataset (CIFAR-10/CIFAR-100) has decreased considerably, which means that the user-specific information cannot be removed without sacrificing its performance and demonstrates its resilience against pruning attack.

\begin{figure}[h]
    \centering
    \includegraphics[width=0.85\linewidth]{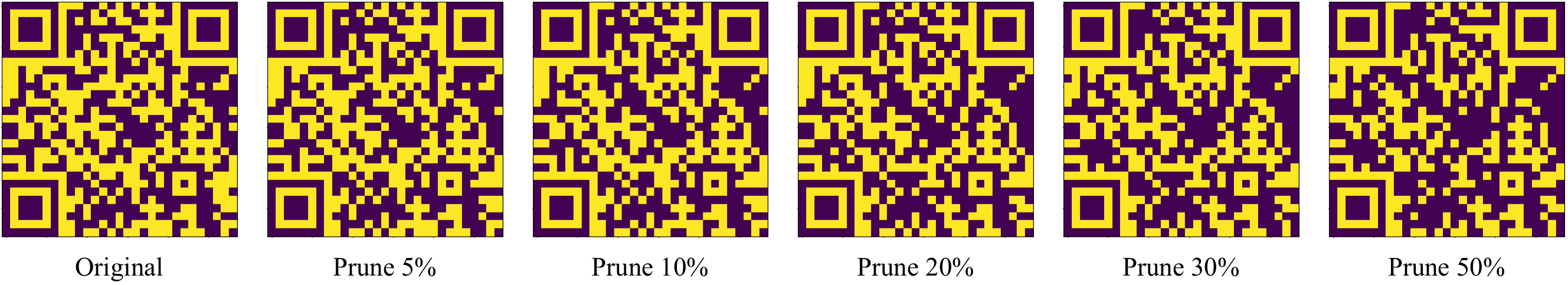}
    \vspace{-1mm}
    \caption{\small QR code extracted from ResNet-20 under pruning attacks with different ratios. The codes extracted under 5\% and 10\% pruning ratio can be easily decoded into readable strings ``signature'', and the code under 20\% pruning ratio can be decoded into ``sigiature'' with tools at \url{https://github.com/merricx/qrazybox/}. } 
    \vspace{-3mm}
    \label{fig:prune_qr_code_layer1}
\end{figure}

\begin{figure}[h]
    \vspace{-1mm}
    \centering
    \includegraphics[width=0.85\linewidth]{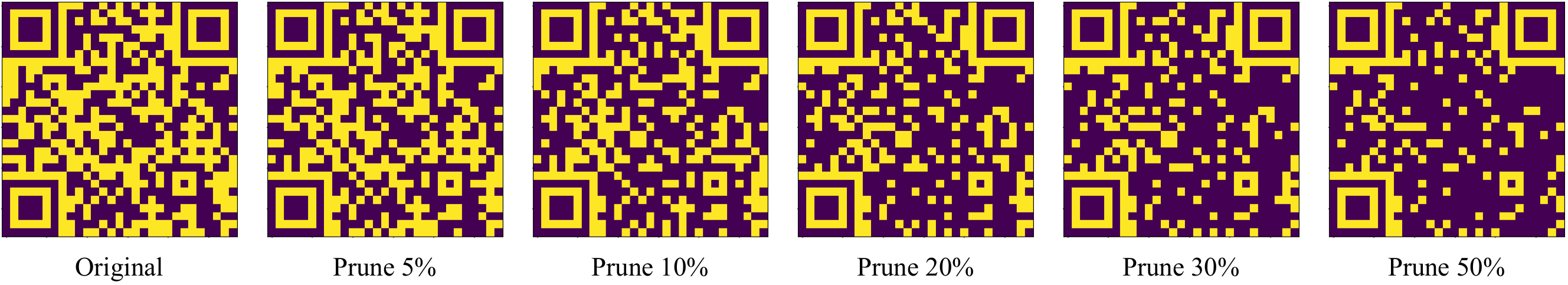}
    \vspace{-1mm}
    \caption{\small QR code extracted from ResNet-18 under pruning attacks with different ratios. The code extracted under 5\% can be easily decoded into a readable string ``signature'', and the code under 10\% pruning ratio can be decoded into ``simçi@5re'' with tools at \url{https://github.com/merricx/qrazybox/}. } 
    \label{fig:prune_qr_code_layer1_res18}
        \vspace{-3mm}
\end{figure}

We then test our proposal for white-box verification (Scheme $\mathcal{V}_2$). In Table~\ref{tab:pruning_trained_scheme_2} we show the inference performance on original datasets after pruning, and also show the QR codes extracted from masks in Figure~\ref{fig:prune_qr_code_layer1}. We can see that the performance of the pruned model will degrade dramatically after 20\% percent of one-shot magnitude pruning and 5\% percent of random pruning. On the contrary, the QR code extracted from the ResNet-20 can be decoded even after 20\% percent of one-shot magnitude pruning. Figure~\ref{fig:prune_qr_code_layer1_res18} shows the QR code extracted from ResNet-18. At the 5\% pruning ratio, the string can be easily decoded into a readable string. At the 10\% pruning ratio, the string can still be partly decoded, although the readability has been reduced. For the pruning ratio greater than 10\%, the inference performance has significantly dropped, making it meaningless to conduct such attack.

%\begin{wraptable}{r}{7cm}
\begin{table}[htbp]
\begin{minipage}[t]{0.47\linewidth}
    \centering
    \vspace{-4mm}
    \caption{\small Inference performance of extremely sparse winning tickets on ResNet-20 and ResNet-18 after model pruning attacks under different pruning methods and pruning ratios. OMP stands for one-shot magnitude pruning. The numbers in brackets stand for the pruning ratios. }
    \resizebox{0.81\linewidth}{!}{
    \begin{tabular}{c|c|c}
    \toprule
    \multirow{2}{*}{Method (Percent)} & \multicolumn{2}{c}{Performance} \\
    \cmidrule{2-3}
    & CIFAR-10 & CIFAR-100 \\
    \toprule
    Scheme $\mathcal{V}_2$ & 91.37\% & 72.35\% \\
    \midrule
    OMP (5\%) & 91.25\%  & 72.27\% \\
    OMP (10\%) & 90.72\% & 71.42\% \\
    OMP (20\%) & 88.03\% & 69.51\% \\
    OMP (30\%) & 80.08\% & 60.31\% \\
    OMP (50\%) & 36.62\% & 9.24\% \\
    \midrule
    Random Pruning (5\%)  & 60.87\% & 58.23\% \\
    Random Pruning (10\%) & 30.49\% & 22.67\%\\
    Random Pruning (20\%) & 11.95\% & 3.23\% \\
    Random Pruning (30\%) & 12.05\% & 1.0\%  \\
    Random Pruning (50\%) & 10.00\% & 1.0\% \\
    \bottomrule
    \end{tabular}}
    %\vspace{0.5em}
    \label{tab:pruning_trained_scheme_2}
    \vspace{-3mm}
\end{minipage}
\hfill%
\begin{minipage}[t]{0.47\linewidth}
\centering
\vspace{-4mm}
    \caption{\small Summary of different types of ambiguity attacks. We show the specification of each attack, \textit{i.e.}, the accessibility of each component to attackers, the attack methods, and the targeted schemes. }
    \resizebox{0.98\textwidth}{!}{
    \begin{tabular}{c|c|c|c}
    \toprule
        Attack name & Attackers can access & How to attack & Attack Scheme \\
        \midrule
        \textit{fake}$_1$ & $\mathbf{W}\odot\mathbf{M}_l$ & Forge $\mathbf{W}\odot\mathbf{M}_s$ & Scheme $\mathcal{V}_1$  \\
        \midrule
        \textit{fake}$_2$ & $\mathbf{W}\odot\mathbf{M}$ & Add noise $\mathbf{W}_\mathrm{noise}\odot\mathbf{M_\mathrm{noise}}$ & Scheme $\mathcal{V}_2$ and $\mathcal{V}_3$  \\
        \textit{fake}$_3$ & $\mathbf{W}\odot\mathbf{M}$ and \texttt{encode}$(\cdot)$ & Replace $\mathbf{M}_s$ & Scheme $\mathcal{V}_2$ and $\mathcal{V}_3$  \\
        \bottomrule
    \end{tabular}}
    \label{tab:type_of_ambiguity_attack}

    \caption{\small Test accuracy and remaining weights after add-on attacks under different rates on ResNet-20, with the matching condition and the decode-ability of the QR code extracted from the masks. }
    \resizebox{0.98\linewidth}{!}{
    \begin{tabular}{c|c|c|c}
    \toprule
    Add-on Rate & Test Accuracy (\% r$_\mathrm{remain}$) & Decode-able? & Match?  \\
    \toprule
    0\%     & 91.53\% (19.369\%) & \ding{51} & \ding{51} \\
    0.5\%   & 91.04\% (19.789\%) & \ding{51} & \ding{51} \\
    1\%     & 90.23\% (20.179\%) & \ding{51} & \ding{55} \\
    2\%     & 86.64\% (21.009\%) & \ding{55} & \ding{55} \\
    5\%     & 79.49\% (23.386\%) & \ding{55} & \ding{55} \\
    10\%    & 71.06\% (27.402\%) & \ding{55} & \ding{55} \\
    \bottomrule
    \end{tabular}}
    \vspace{-3mm}
    \label{tab:add_on_qr_code}
    
\end{minipage}

%\end{wraptable}
\end{table}
%\begin{table}[htbp]
%    \centering
%    \caption{Inference performance of extremely sparse winning tickets on ResNet-20 and ResNet-18 after model pruning attacks under different pruning methods and pruning ratios. OMP stands for one-shot magnitude pruning. The numbers in brackets stand for the pruning ratios. }
%    \begin{tabular}{c|c|c|c|c}
%    \toprule
%    \multirow{2}{*}{Pruning Method (Percent)} & \multicolumn{2}{c|}{ResNet-20} & \multicolumn{2}{c}{ResNet-18} \\
%    \cmidrule{2-5}
%    & CIFAR-10 & CIFAR-100 & CIFAR-10 & CIFAR-100 \\
%    \toprule
%    Baseline & 91.37\% & 66.14\% & 93.56\% \\
%    \midrule
%    OMP (5\%) & 91.25\% & 65.12\% & 93.57\% \\
%    OMP (10\%) & 90.72\% & 62.00\% & 93.55\% \\
%    OMP (20\%) & 88.03\% & 48.34\% & 93.14\% \\
%    OMP (30\%) & 80.08\% & 32.35\% & 90.95\% \\
%    OMP (50\%) & 36.62\% & 7.14\% & 33.31\% \\
%    \midrule
%    Random Pruning (5\%)  & 60.87\% & 17.79\% & 84.23\%\\
%    Random Pruning (10\%) & 30.49\% & 6.45\% & 67.86\% \\
%    Random Pruning (20\%) & 11.95\% & 1.90\% & 14.19\% \\
%    Random Pruning (30\%) & 12.05\% & 1.52\% & 10.00\%\\
%    Random Pruning (50\%) & 10.00\% & 1.00\% & 10.00\% \\
%    \bottomrule
%    \end{tabular}
%    \vspace{0.5em}
%    \label{tab:pruning_trained_scheme_2}
%\end{table}

%It can be clearly seen from Figure~\ref{tab:pruning_cifar} and \ref{tab:pruning_cifar100} that the trained model cannot be pruned, so that all the schemes are robust to model pruning attacks. The performance drops are significantly even pruning only 1\% of the trained weights, showing the effectiveness and sensitivity of our protection schemes.  

\subsection{Resilience Against Ambiguity Attacks}
\vspace{-2mm}
In this section, we will evaluate the robustness against ambiguity attacks summarized in Table~\ref{tab:type_of_ambiguity_attack}.

\vspace{-2mm}
\paragraph{Scheme $\mathcal{V}_1$}

\begin{figure}[htbp]
    \centering
        \vspace{-1mm}
    \includegraphics[width=0.9\linewidth]{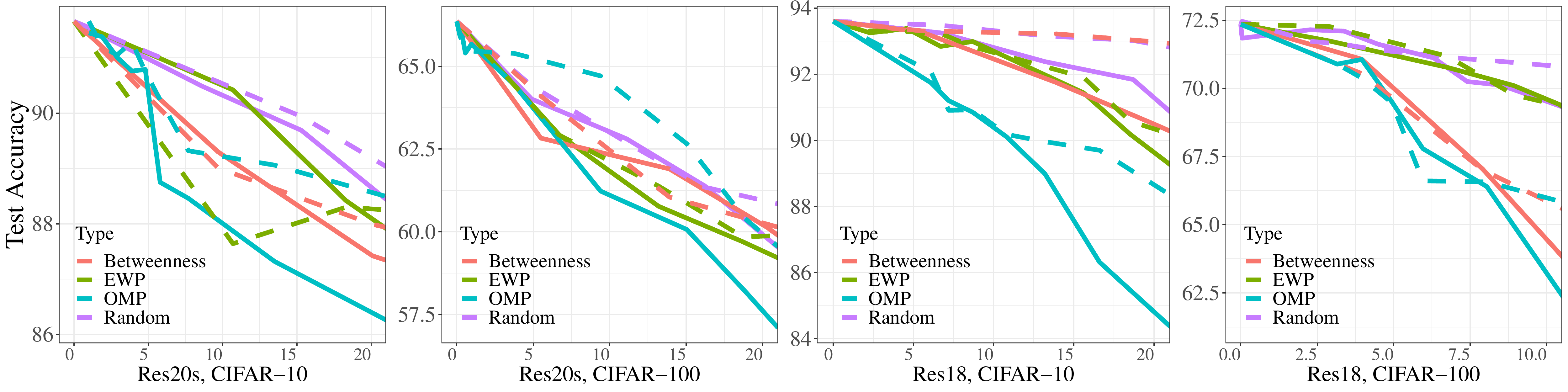}
        \vspace{-1mm}
    \caption{\small Random attacks on Scheme $\mathcal{V}_1$. The $x$-axis is the relative sparsity of the key masks. The solid/dashed lines represent the performance \textbf{before}/\textbf{after} random attacks.}
        \vspace{-3mm}
    \label{fig:random_attacks_scheme_1}
\end{figure}

($\textit{fake}_1$: Attackers can access $\mathbf{W}\odot\mathbf{M}_l$ only)
The goal of $\textit{fake}_1$ is to forge a new key mask $\mathbf{M}_s'$ with new underlying weights $\mathbf{W}'$. As the attacker has no prior information on $(\mathbf{W}\odot\mathbf{M}_s)$, the forging process can only be performed randomly. From Figure~\ref{fig:random_attacks_scheme_1} we can see that such an attack method is not practical as the performance gap is much greater than $\epsilon_f (= 1\%)$ after using random key masks. For example, if we adopt OMP as the scoring function to construct the key masks, we only need a key mask of around 10\% relatively sparsity to make the model resistant to random attacks.   

%There are two ways to forge such set: one random method and one reverse-engineer method. Random method refers to the setting that the attackers has no prior information on how the key paths are generated. The reverse-engineer version means that the attackers know the algorithm for generating the paths. 

%The results are shown in Figure~\ref{fig:random_attacks_scheme_1}. 

\begin{figure}[h]
    \centering
        \vspace{-1mm}
    \includegraphics[width=0.85\linewidth]{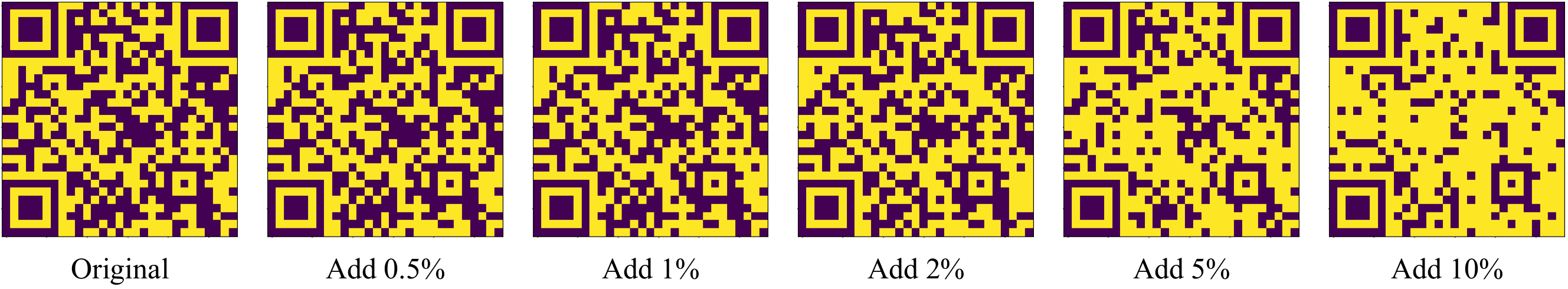}
        \vspace{-1mm}
    \caption{\small Visualization of QR code extracted and processed under different add-on rates. }    
    \vspace{-3mm}
    \label{fig:add_on_qr_code}
\end{figure}

\paragraph{Scheme $\mathcal{V}_2$ and $\mathcal{V}_3$}
($\textit{fake}_2$: Attackers can access $\mathbf{W}\odot\mathbf{M}$ but not knowing $\texttt{encode}(\cdot)$)
One might use add-on attacks and try to ``contaminate'' the information we embed in the sparse mask. Specifically, we randomly add noises to the position where the weights are pruned. We test with add-on rates ranging from 0\% to 10\% since a 10\% efficiency gap will diminish the value of attacking the model. The results are shown in Table~\ref{tab:add_on_qr_code} and Figure~\ref{fig:add_on_qr_code}. From the table, we can see that introducing 1\% of noise to the trained model will un-match the attacked model (\textit{i.e.,} the performance gap becomes greater than 1\%). For the add-on rates smaller than 1\%, the QR code embedded in the sparse mask can be normally decoded into a normal string. Such results prove that both Scheme $\mathcal{V}_2$ and $\mathcal{V}_3$ are resistant to attack $\textit{fake}_2$. 
%The attackers might prune the model in order to remove the QR code from the models. However, our method show great resistance towards add-on attacks. The QR codes extracted after pruning attacks are shown in Figure~\ref{fig:add_on_qr_code}. The correction rate of QR code under different add-on rates are shown in Table~\ref{tab:add_on_qr_code}. 

($\textit{fake}_3$: Attackers can access $\mathbf{W}\odot\mathbf{M}$ and $\texttt{encode}(\cdot)$) If the attacker knows about the \texttt{encode} function for generating the $\mathbf{M}_s$, a similar but fake signature mask $\mathbf{M}_s'$ which contains a different signature can be generated in the same way. However, as shown in Figure~\ref{fig:tesear}, without the finder, alignment, and version patterns, one can hardly tell which part belongs to a QR code. %Therefore, one can only randomly choose a region in the trained weights, and replace it with $\mathbf{M}_s' \odot \mathbf{W}'$, where $\mathbf{W}$ has the same shape as the region to attack but whose values are randomly drawn. 
Even if the attacker knows the position where the code is embedded (namely an insider attack~\cite{fan2019rethinking}), directly replacing the embedded region with a new signature mask $\mathbf{M}_s'$ and $\mathbf{W}'$ (noise) will also considerably degrade the performance of the attacked model since a large amount of ``incorrect'' weights are introduced. For example, for ResNet-20 on CIFAR-10, the test accuracy of the attacked model will drop from $91.37\%$ to $57.00\%$, which is nearly a 50\% degradation in performance. Such a big loss shows that it is infeasible to perform the insider attack. 

\vspace{-0.8em}
\section{Conclusion and Discussion of Broad Impact} \label{sec:impact}
\vspace{-0.7em}
\noindent LTH offers superior sparse models through burdensome explorations, serving as an intriguing yet expansive solution for resource-constrained applications. It motivates the necessity of protecting the copyright of these precious winning tickets. We investigate a brand new verification technique by leveraging the sparse structural information, which embeds signatures into lottery tickets' typologies. Extensive results verify our proposal's effectiveness and robustness against diverse malicious attacks.

This work is scientific in nature and should bring positive societal impacts. Note that every second, giant and start-up companies have invested billions of dollars to identify superior yet light-weight compact deep neural networks virtually. We believe our new \textit{lottery verification} mechanism can assist both industry and academia in defending their interests from illegal distribution or usage.

\bibliographystyle{unsrt}
\bibliography{ref}

\clearpage

\appendix

\renewcommand{\thepage}{A\arabic{page}}  
\renewcommand{\thesection}{A\arabic{section}}   
\renewcommand{\thetable}{A\arabic{table}}   
\renewcommand{\thefigure}{A\arabic{figure}}

\section*{Checklist}

%%% BEGIN INSTRUCTIONS %%%
% The checklist follows the references.  Please
% read the checklist guidelines carefully for information on how to answer these
% questions.  For each question, change the default \answerTODO{} to \answerYes{},
% \answerNo{}, or \answerNA{}.  You are strongly encouraged to include a {\bf
% justification to your answer}, either by referencing the appropriate section of
% your paper or providing a brief inline description.  For example:
% \begin{itemize}
%   \item Did you include the license to the code and datasets? \answerYes{See Section.}
%   \item Did you include the license to the code and datasets? \answerNo{The code and the data are proprietary.}
%   \item Did you include the license to the code and datasets? \answerNA{}
% \end{itemize}
% Please do not modify the questions and only use the provided macros for your
% answers.  Note that the Checklist section does not count towards the page
% limit.  In your paper, please delete this instructions block and only keep the
% Checklist section heading above along with the questions/answers below.
%%% END INSTRUCTIONS %%%

\begin{enumerate}

\item For all authors...
\begin{enumerate}
  \item Do the main claims made in the abstract and introduction accurately reflect the paper's contributions and scope?
    \answerYes{}
  \item Did you describe the limitations of your work?
    \answerYes{Please refer to Section~\ref{sec:impact}.} 
  \item Did you discuss any potential negative societal impacts of your work?
    \answerYes{Please refer to Section~\ref{sec:impact}.} 
  \item Have you read the ethics review guidelines and ensured that your paper conforms to them?
    \answerYes{}
\end{enumerate}

\item If you are including theoretical results...
\begin{enumerate}
  \item Did you state the full set of assumptions of all theoretical results?
    \answerNA{Our work completely focuses on empirical investigation.}
	\item Did you include complete proofs of all theoretical results?
    \answerNA{Our work completely focuses on empirical investigation.}
\end{enumerate}

\item If you ran experiments...
\begin{enumerate}
  \item Did you include the code, data, and instructions needed to reproduce the main experimental results (either in the supplemental material or as a URL)?
    \answerYes{We use the publicly available datasets (section~\ref{sec:setting}) and attach our codes with instructions to the supplement for better reproducibility.}
  \item Did you specify all the training details (e.g., data splits, hyperparameters, how they were chosen)?
    \answerYes{We detail our experiment settings in section~\ref{sec:setting}}
	\item Did you report error bars (e.g., with respect to the random seed after running experiments multiple times)?
    \answerNo{We conducted a single run for each experiment due to the limited resources. We will repeat experiments and report error bars in the future.}
	\item Did you include the total amount of compute and the type of resources used (e.g., type of GPUs, internal cluster, or cloud provider)?
    \answerYes{Please refer to section~\ref{sec:setting}}
\end{enumerate}

\item If you are using existing assets (e.g., code, data, models) or curating/releasing new assets...
\begin{enumerate}
  \item If your work uses existing assets, did you cite the creators?
    \answerYes{As shown in section~\ref{sec:setting}, we use publicly available datasets and cite the creators.}
  \item Did you mention the license of the assets?
    \answerNo{The licenses of used datasets are provided in the cited paper.}
  \item Did you include any new assets either in the supplemental material or as a URL?
    \answerYes{All used datasets are publicly available, and all our codes are provided at \url{https://github.com/VITA-Group/NO-stealing-LTH}.}
  \item Did you discuss whether and how consent was obtained from people whose data you're using/curating?
    \answerNA{We did not use/curate new data.}
  \item Did you discuss whether the data you are using/curating contains personally identifiable information or offensive content?
    \answerNA{All adopted datasets are publicly available, and we believe there are no issues of personally identifiable information or offensive content.}
\end{enumerate}

\item If you used crowdsourcing or conducted research with human subjects...
\begin{enumerate}
  \item Did you include the full text of instructions given to participants and screenshots, if applicable?
    \answerNA{}
  \item Did you describe any potential participant risks, with links to Institutional Review Board (IRB) approvals, if applicable?
    \answerNA{}
  \item Did you include the estimated hourly wage paid to participants and the total amount spent on participant compensation?
    \answerNA{}
\end{enumerate}

\end{enumerate}

\section{More Methodology Details} \label{sec:more_methods}

\paragraph{More of ownership verification schemes.} Table~\ref{tab:my_label} summarizes our proposed ownership verification regimes. There are five different phases in each of our schemes: 1) \textit{Ticket finding}: finding the extremely sparse winning tickets. Multiple rounds of the train-prune-retrain process are involved in this phase for finding the extremely sparse winning tickets; 2) \textit{Pre-Process}: pre-process the extremely sparse winning ticket for applying each scheme. For example, we need to construct the key masks if using the Scheme $\mathcal{V}_1$; 3) \textit{Re-training}: this process is unique for the winning tickets that we will train the extremely sparse winning ticket again to match the performance of the dense model; 4) \textit{Inference}: the inference process is to perform an inference process on the test dataset; 5) \textit{Validation}: This process is to validate the ownership of the (trained/untrained) extremely sparse winning ticket. 

\begin{table}[htbp]
\vspace{-0.5em}
    \centering
    \caption{Summary of different ownership verification schemes. The re-training phase can be either done by the ticket owner or the legitimate users. }
    \resizebox{0.95\textwidth}{!}{
    \begin{tabular}{c|c|c|c}
    \toprule
         & Scheme $\mathcal{V}_1$ & Scheme $\mathcal{V}_2$ & Scheme $\mathcal{V}_3$ \\
         \midrule
        Ticket Finding & No additional technique & No additional technique & No additional technique \\
        \midrule
        \multirow{2}{*}{Pre-Process} & Split key masks and locked masks & Calculate $\mathbf{M}_s$ using \texttt{encode}$(\cdot)$ & Calculate $\mathbf{M}_s$ using \texttt{encode}$(\cdot)$ \\
        & Distribute both the masks & Embed $\mathbf{M}_s$ into $\mathbf{M}$ and distribute & Embed $\mathbf{M}_s$ into $\mathbf{M}$ and distribute \\
         \midrule
        \multirow{3}{*}{Re-training} & \multirow{3}{*}{Recover the masks} & \multirow{3}{*}{No additional technique} & \multirow{3}{*}{Training with the trigger set $\mathbf{T}$} \\
        &  &  &  \\ 
        &  &  &  \\
        \midrule
        \multirow{2}{*}{Inference} &  Keys masks are required & \multirow{2}{*}{No additional technique}  & \multirow{2}{*}{No additional technique} \\
        & Slight overhead for recovering the masks & & \\
        \midrule
        \multirow{2}{*}{Validation} & \multirow{2}{*}{Auto-verified by performance} & \multirow{2}{*}{Extract $\mathbf{M}_s$ and decode} & Extract $\mathbf{M}_s$ and decode \\
        & & & Inference on trigger set $\mathbf{T}$ \\
        \bottomrule
    \end{tabular}}
    \vspace{-0.5em}
    \label{tab:my_label}
\end{table}

% \vspace{-1em}
\section{More Experimental Results} \label{sec:more_results}

\paragraph{Extremely sparse winning tickets on ResNet-50.}
%The extremely sparse winning tickets on ResNet-50
On CIFAR-10, the remaining weights of the extremely sparse winning ticket is 13.19\% (pruning specification: (7,1,6,0)) while the performance is 94.38\% (0.04\% drop). On CIFAR-100, the proportion of remaining weights of the extremely sparse winning ticket is 43.926\% (pruning specification: (2,3,0,6)) while the performance is 75.84\% (0.03\% drop). On ImageNet, the proportion of remaining weights of the extremely sparse winning ticket is 16.97\%, and the performance is 75.97\% (0.01\% higher). 

\paragraph{Extremely sparse winning tickets on VGG-16. }
On CIFAR-10, the proportion of the remaining weights of the extremely sparse winning ticket is 1.44\%, while the performance is 93.10\% (0.04\% higher). On Tiny-ImageNet, the proportion of the remaining weights of the extremely sparse winning ticket is 6.81\%, while the performance is 58.12\% (0.19\% higher). 

\begin{figure}[h]
    \centering
    \vspace{-2mm}
    \includegraphics[width=0.9\linewidth]{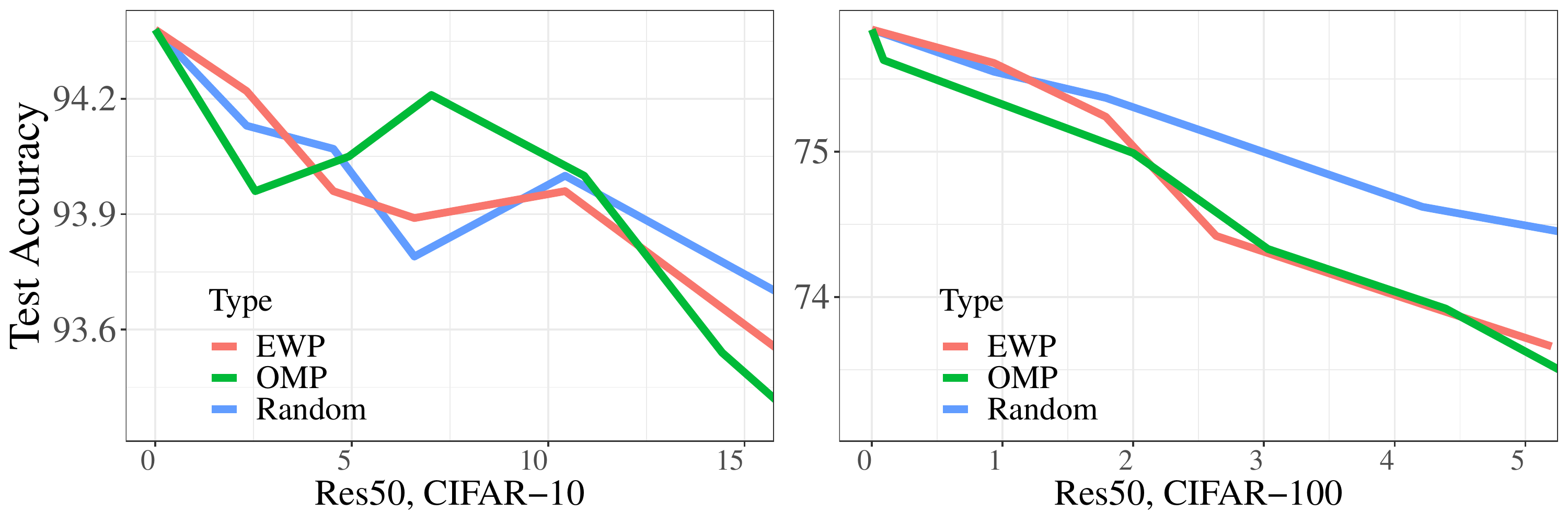}
    \vspace{-3mm}
    \caption{\small Effectiveness of Scheme $\mathcal{V}_1$: Re-training without key masks generated by three methods: EWP, OMP, Random. The $x$-axis is the relative sparsity w.r.t the extreme ticket.}
    \label{fig:scheme1_on_res50}
    \vspace{-4mm}
\end{figure}

\paragraph{Scheme $\mathcal{V}_1$ on ResNet-50.} Figure~\ref{fig:scheme1_on_res50} shows the results of retraining the extremely sparse winning tickets without key masks. Multiple scoring functions (OMP, EWP, Random) are explored. It can be seen from the graph that on CIFAR-10, we need key masks with an approximately 15\% relative sparsity to create a 1\% performance gap, while on CIFAR-100, we need key masks with a relative sparsity of 5\% approximately. ResNet-50 has greater model capacity than ResNet-20 and ResNet-18, so it is reasonable that we need more elements removed to reduce the performance significantly. 

On ImageNet, the performance of the retrained model is 75.39\% when the relative sparsity is 0.4\%, and the performance is 72.88\%, which is nearly 3 percent lower when the relative sparsity is 5\%. It proves that our Scheme $\mathcal{V}_1$ can work on large-scale datasets. 

\begin{figure}[h]
    \centering
    \vspace{-2mm}
    \includegraphics[width=0.9\linewidth]{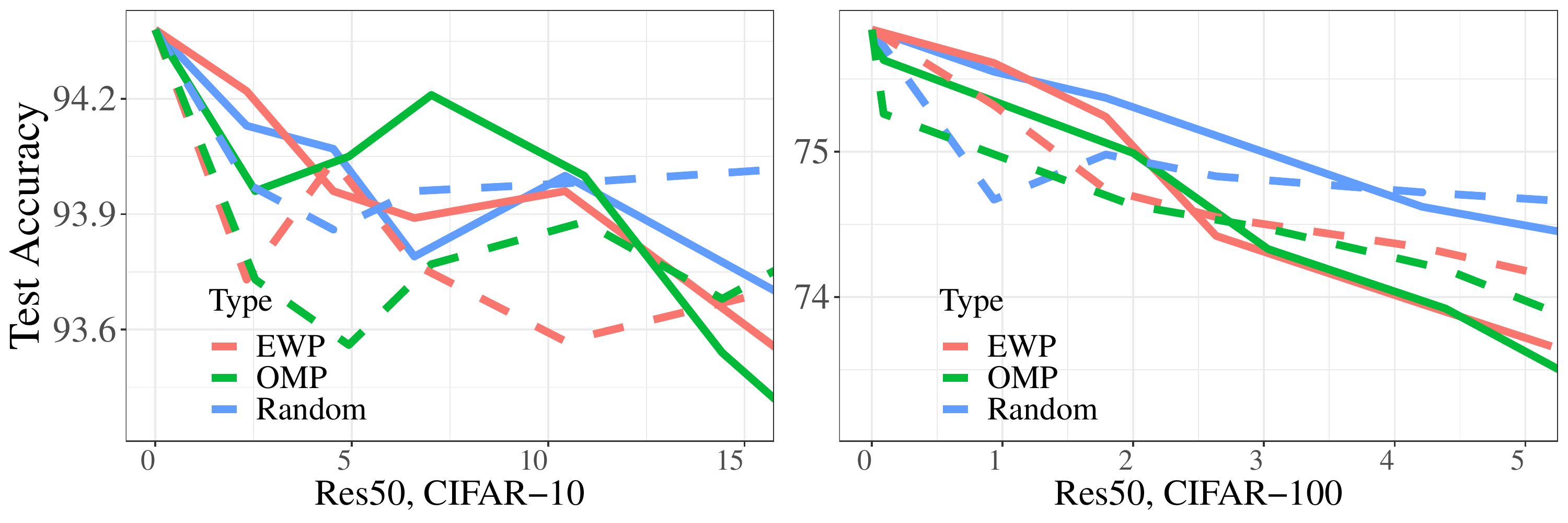}
    \vspace{-3mm}
    \caption{\small Random attacks on Scheme $\mathcal{V}_1$ on ResNet-50. The $x$-axis is the relative sparsity of the key masks. The solid/dashed lines represent the performance \textbf{before}/\textbf{after} random attacks.}
    \label{fig:scheme1_on_res50_1}
    \vspace{-4mm}
\end{figure}

\paragraph{Random ambiguity attacks on ResNet-50 under scheme $\mathcal{V}_1$.}
Figure~\ref{fig:scheme1_on_res50_1} shows the results of using random key masks for retraining the extremely sparse winning ticket for ResNet-50 on CIFAR-10 and CIFAR-100. It can be clearly seen from the graph that the random key masks will not contribute to recovering the performance of the trained model and even harm the test accuracy under some circumstances. On ImageNet, the accuracy of recovering masks with random connections is 75.32\% and 74.57\% when the relative sparsity is 0.4\% and 5\%, respectively. The performance gaps, which can be seen easily from the graphs and numbers, have demonstrated the robustness of Scheme $\mathcal{V}_1$ against the ambiguity attack.

\paragraph{Scheme $\mathcal{V}_1$ on VGG-16.}
On CIFAR-10, the performance of the retrained model without key masks is 88.63\% when the relative sparsity of the key masks is 8\%, and the performance after recovering with random connections is only 91.96\%. On Tiny-ImageNet, the performance of the retrained model without key masks/with random key masks is 48.97\%/52.86\%. These results show the effectiveness and robustness of our Scheme $\mathcal{V}_1$. 

\paragraph{Scheme $\mathcal{V}_2$ and  $\mathcal{V}_3$  on VGG-16.}
We further examine the effectiveness and the robustness of the Scheme $\mathcal{V}_2$ and $\mathcal{V}_3$. The QR code embedded we put in the sparse mask of VGG-16 can still be partly decoded when the pruning ratio is 10\%, while the test accuracy is 57.26\% after pruning (0.7\% lower) on Tiny-ImageNet. As for the Scheme $\mathcal{V}_3$, the test accuracy on Tiny-ImageNet decreases to 56.44\% (over 1.5\% lower) after pruning 20\% of the trained model while the test accuracy on the trigger set is still 100\%. All these phenomena show the effectiveness and robustness of our Scheme $\mathcal{V}_2$ and $\mathcal{V}_3$ on VGG-16.

\end{document}